\documentclass{article}
\usepackage{spconf,amsmath,graphicx}
\usepackage{graphicx}
\usepackage{multirow}

\makeatletter
\renewcommand*{\@fnsymbol}[1]{\dagger}
\makeatother

\def\L{{\cal L}}

\title{A UNIFIED SEQUENCE-TO-SEQUENCE FRONT-END MODEL FOR MANDARIN TEXT-TO-SPEECH SYNTHESIS}

\name{Junjie Pan$^{\star}$, Xiang Yin$^{\star}$, Zhiling Zhang\sthanks{work done while doing an internship at Bytedance.}, Shichao Liu$^{\star}$, Yang Zhang$^{\star}$, Zejun Ma$^{\star}$, Yuxuan Wang$^{\star}$}

\address{$^{\star}$Bytedance AI-Lab\\$^{\dagger}$Shanghai Jiaotong University\\\tt\small{panjunjie.jeff@bytedance.com}}

\begin{document}
%
\maketitle
\begin{abstract}
In Mandarin text-to-speech (TTS) system, the front-end text processing module significantly influences the intelligibility and naturalness of synthesized speech. Building a typical pipeline-based front-end which consists of multiple individual components requires extensive efforts. In this paper, we proposed a unified sequence-to-sequence front-end model for Mandarin TTS that converts raw texts to linguistic features directly. Compared to the pipeline-based front-end, our unified front-end can achieve comparable performance in polyphone disambiguation and prosody word prediction, and improve intonation phrase prediction by 0.0738 in F1 score. We also implemented the unified front-end with Tacotron and WaveRNN to build a Mandarin TTS system. The synthesized speech by that got a comparable MOS (4.38) with the pipeline-based front-end (4.37) and close to human recordings (4.49).
\end{abstract}
\section{Introduction}
\label{sec:introduction}
In Mandarin text-to-speech (TTS) synthesis, the front-end casts great influence on the  intelligibility and naturalness of synthesized speech. It extracts various linguistic features from the raw text, aiming to provide enough information for the back-end to synthesize close-to-human speech.

The typical Mandarin TTS front-end is a pipeline-based system, consist of a series of text processing components, such as text normalization (TN), Chinese word segmentation (CWS), part-of-speech (POS) tagging, grapheme-to-phoneme (G2P) conversion, and prosody prediction. This structure enables us to divide and conquer the complicated front-end task. However, this serial structure also brings several problems. One is error propagation, that small errors in single component flow to its downstream components, and can severely damage final results way beyond just adding up the individual errors. The second is the complex feature engineering and data labeling work, as each component requires different input features and output labels. Another is that front-end components need to be trained and optimized separately, which causes a misalignment in optimization and make the whole training procedure much complicated.

To address the issues above, we proposed a unified sequence-to-sequence front-end structure, which predict phoneme, tone, and prosody sequences from the raw text directly. Compared to the pipeline-based front-end, our unified front-end could achieve comparable performance in polyphone disambiguation and prosody word prediction, and boost intonation prosody prediction by 0.0738. When implemented with Tacotron~\cite{wang2017tacotron} and WaveRNN~\cite{kalchbrenner2018efficient} in a Mandarin TTS system, the synthesized speech by the unified front-end got a 4.38 mean opinion score (MOS), which is comparable with the pipeline-based front-end (4.37) and close to the human recordings (4.49). The main contribution of this work is to provide a unified  front-end structure to build a high-quality Mandarin TTS system more quickly and easily.


\section{Background}
\label{sec:background}
In this section, we briefly review some major components in Mandarin TTS front-end, including TN, CWS, POS tagging, G2P, and prosody prediction.
\subsection{Text Normalization}
\label{subsec:tn}
Text normalization (TN) is the process of converting non-standard words into spoken-form words with unambiguous pronunciation. Traditional TN is rule-based, but the TN applied in this work is a novel hybrid method~\cite{zhang2020nntn} and the same for all experiments.

\subsection{CWS and POS tagging}
\label{subsec:cws_pos}
Unlike the English language, Mandarin is a character-based language, so we need CWS to extract word boundaries from the raw text. Accurate word boundaries are the key to build high-precision POS, G2P, and prosody prediction models, and CWS errors are likely to decrease the intelligibility of the synthesized speech. However, the ambiguity resolution in CWS is always a controversial issue~\cite{teahan2000compression}. POS tagging takes an important role in polyphone disambiguation, and inaccurate POS usually results in incorrect polyphone pronunciations. The outputs of CWS and POS are not used directly by the front-end in Mandarin TTS, but they are crucial auxiliary features in G2P and prosody prediction tasks. Therefore, we add an auxiliary module to extract latent representations of CWS and POS in our proposed model.

\subsection{Prosody Prediction}
\label{subsec:pwpp}
The prosody structure in Mandarin TTS determines the naturalness to a great extent, which has been proved in previous researches~\cite{ying2001rnn, lu2019implementing}. It is usually divided into 3 levels (from low to high) - prosody word (PW), prosody phrase (PP), and intonation phrase (IP), where the higher level is based on the lower level. Typical prosody prediction methods include rule-based models and statistical models like CRF~\cite{qian2010automatic}, and RNN~\cite{ying2001rnn}. Recently, Multi-Task Learning (MTL) architecture is also applied in prosody prediction~\cite{pan2019mandarin}. In our proposed model, we predict prosody labels with phoneme and tone labels together within a joint modeling structure.

\subsection{Grapheneme-to-Phoneme}
\label{subsec:g2p}
As described in ~\cite{xu2004grapheme}, the main objective in Mandarin G2P is polyphone disambiguation(PD), which affect the speech quality significantly in TTS. Additionally, tone sandhi and Erhua are also the key issues in the intelligibility of Mandarin TTS. The most popular method for PD is to apply the ME model for each polyphone~\cite{mao2007inequality}. Using a unified model for all polyphones were also investigated~\cite{huang2008disambiguating}, recently. However, all these methods can only process one polyphone at a time, and need post-fixing rules for tone sandhi and Erhua. In our proposed model, the PD, tone sandhi, and Erhua issues in a sequence of text can be resolved at once.

\section{Model Architecture}
\label{sec:model_architecture}
In this section, we proposed a unified sequence-to-sequence front-end model, to generate phoneme, tone, and prosody sequences from raw text inputs directly. It helps to simplify the Mandarin TTS front-end pipeline and avoids complicated post-fixing rules in tone sandhi and Erhua. The idea was derived from ~\cite{vaswani2017attention, wang2017tacotron, skerry2018towards, shen2018natural}.

\subsection{Data flow}
\label{subsec:dataflow}
\begin{figure}
    \centering
    \includegraphics[width=\columnwidth]{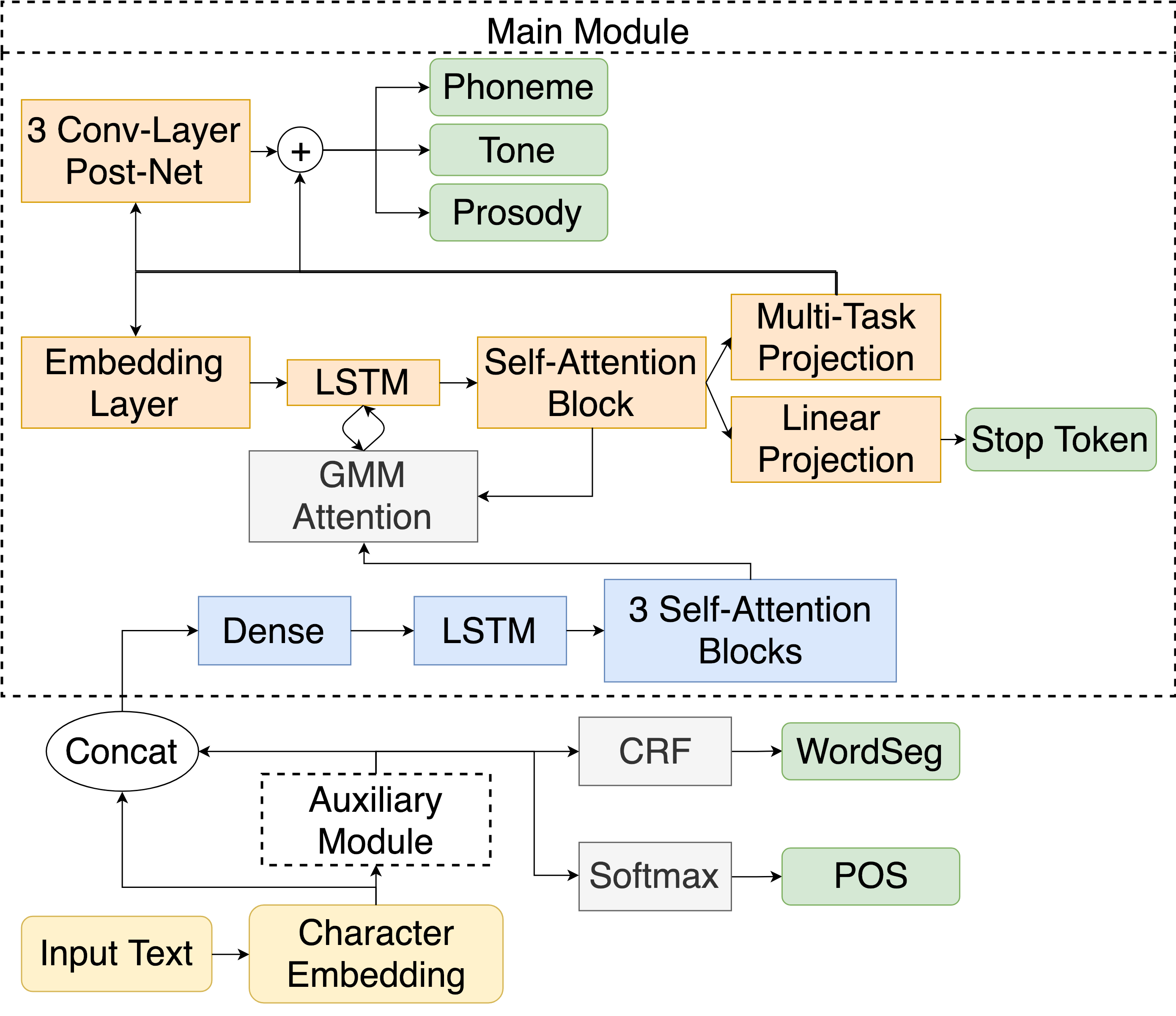}
    \caption{Structure of the unified sequence-to-sequence front-end}
    \label{fig:shooter}
\end{figure}

As shown in Figure \ref{fig:shooter}, the raw input text is first converted to a sequence of 300-dimension vectors by a pre-trained character embedding mapping. The character embedding mapping is generated from a Word2Vec~\cite{mikolov2013distributed} model, which was trained by a 1Gb Chinese corpus from Wikipedia in character level. The character embedding sequence is then passed to an auxiliary module, and its outputs are one-hot vectors for CWS and POS tags, where CWS has 4 tags (B, M, E, S) and POS has 99 tags.

The dense presentation before the CRF/Softmax layer in the auxiliary module is concatenated with the original character embedding as the inputs for the main module. Compared to the classification outputs from the auxiliary module, this representation can mitigate the error-propagation issue, and makes our model more robust.

Outputs of the main module are phoneme, tone, and prosody sequences, which are used as the linguistic features for the back-end in Mandarin TTS. A stop token is also generated here, to terminate the decoder prediction in inference.

\subsection{Auxiliary Module}
\label{subsec:auxiliary_module}
The auxiliary module acts as a feature extractor to extract CWS and POS representations from raw text inputs, which are expected to provide more information for the main module. Inspired by~\cite{shaheen2016impact, uszkoreit2017transformer}, we proposed two auxiliary module structures - the dilated-CNN (DCNN) and the transformer encoder (TE). The DCNN model consists of 3 dilated CNN layers, with kernel size=5, filter number=128, and dilation rate=1,2,4 respectively. The TE model consists of a 256-unit LSTM layer and 1 self-attention block with 8 heads. Positional embedding is applied to the TE as well.

\subsection{Main Module}
\label{subsec:main_module}
The main module is a encoder-decoder structure with Gaussian Mixture Model(GMM) attention~\cite{wang2018style}. In the encoder part, there is a 128-unit LSTM layer, followed by a 32-unit projection layer. The outputs of the projection layer are fed into 3 blocks of self-attention with 8 heads. The encoder converts the concatenation of the character embedding and dense CWS\&POS representations to hidden feature representations, which the decoder consumes to predict the phoneme, tone, and prosody labels.

In the decoder part, there is a 1024-unit LSTM layer with 1 self-attention block with 8 heads. During the inference stage, semi-auto-regressive (SAR) method is applied in each decoding step, to improve decoder's performance. With the inspiration of~\cite{shen2018natural}, the full decoded output sequence is then passed into a post-net, which comprises 3 CNN layers with kernel size=5 and filter number=128. The post-net generates a residual, which is added to decoder's predictions, to promote the final performance. Joint modeling is applied in this part, to generate phoneme, tone, and prosody sequences together.

We used the GMM attention rather than local sensitive attention in our model, as it shows better monotonicity and alignments between the inputs and outputs. Different from~\cite{wang2018style}, we replace exponential activation with softplus activation in GMM attention, which is expected to increase the stability substantially during training~\cite{battenberg2019effective}.


\subsection{Semi Auto Regressive (SAR) Inference}
\label{subsec:semi_teacher_forcing}
In the inference stage, the common method is to use the auto-regressive (AR) method to predict results at each step. However, in our task, the phoneme only needs to be predicted for polyphones and can be identified for non-polyphones in advance. In this case, we proposed a semi-auto-regressive (SAR) method to improve the decoding performance, where the correct phoneme replaces the predicted phoneme at every non-polyphone step. A decoder mask is generated from the raw input text to tell the decoder helper whether the current step is for polyphones or not. The prediction for tones and prosody labels keeps AR behavior.

\section{Experiments and Results}
\label{sec:exp_and_res}
\subsection{Dataset}
\label{subsec:dataset}
We use the following datasets:
\begin{description}
    \item[People's Daily Corpus:] 940,000 Mandarin utterances with word boundaries and POS tags, generated from People's Daily (2004) corpus, released by Peking University. It is used for auxiliary-module training, validation, and test with the ratio of 8:1:1.

    \item[Internal Corpus:] Two groups of Mandarin corpus with prosody intervals and Pinyin labels. One contains 117,967 utterances for the unified front-end training and validation with the ratio of 9:1. The other is the golden test set with 49,466 utterances.


\end{description}

\subsection{Training and Test}

Due to the post-net structure described in Section~\ref{subsec:main_module}, we optimize two group of losses. One is the before-post-net loss, consist of 4 cross-entropy (CE) losses. The other is the after-post-net loss, consists of 3 negative log-likelihoods (LL). The final loss function becomes:
\setlength\arraycolsep{2pt}
\begin{eqnarray}
loss & = & \sum_{m,n,k,s\in\mathcal{D}}\{CE(\tilde{m}_{before}, \hat{m}) + CE(\tilde{n}_{before}, \hat{n}) \nonumber\\
&& + CE(\tilde{k}_{before}, \hat{k}) + CE(\tilde{s}, \hat{s}) + \mathcal{LL}(\tilde{m}_{after}, \hat{m})\nonumber\\
&& + \mathcal{LL}(\tilde{n}_{after}, \hat{n}) + \mathcal{LL}(\tilde{k}_{after}, \hat{k})\}
\end{eqnarray}
where $\mathcal{D}$ represents the training corpus and $\mathcal{M}, \mathcal{N}, \mathcal{K}, \mathcal{S}$ represent phoneme, tone, prosody sequences, and the stop token respectively.

We first train the auxiliary module with 80\% of the People's Daily corpus, and then fine-tune the whole unified front-end with 90\% of the internal corpus. In the fine-tuning stage, we apply the scheduled sampling method to increase its robustness to the wrong prediction result in the previous time-step, as our experiments showed that the teacher-forcing training would cause the unified front-end to generate unacceptable results in the test.

In the test stage, we used 10\% of the People's Daily corpus for the auxiliary module and the golden test set from the internal corpus for the unified front-end. Our proposed semi-auto-regressive method was applied in this stage.

\subsubsection{System Configuration}
\label{subsec:model_settings}
When training the auxiliary module, we set the dropout rate to 0.1, and apply the label smoothing method with ratio=0.1. The batch bucketing method is used to speed up the training process, and 11 buckets are set with an upper boundary of 200 for input texts.

When fine-tuning the unified front-end, the same dropout and label smoothing settings are set as the auxiliary module. Batch bucketing is applied with 13 buckets and an upper boundary of 90, which aligns with the setting used in our Tacotron training. Scheduled sampling is implemented in the decoder, which starts at 20,000 step with teacher forcing ratio decaying from 1 to 0 in 50,000 steps . The start point 20,000 is a empirical number for our front-end to converge under teacher forcing.

\subsection{Results and Analysis}
\subsubsection{Selection of Auxiliary Module}
\label{subsubsec:auxiliary_module}
We carried out experiments on two auxiliary module structures with different outputs. From Table~\ref{tab:auxiliary_module} it can be found that the DCNN got the highest F1 score in the single CWS task, and the TE performed better in the single POS task and the CWS + POS task. These 3 settings were implemented in our unified front-end in the following experiment.

\begin{table}
\centering
\resizebox{\columnwidth}{!}{%
\begin{tabular}{c|cc|cc|cl}
\hline\hline
\textbf{Task}    & \multicolumn{2}{c|}{\textbf{CWS}} & \multicolumn{2}{c|}{\textbf{POS}} & \multicolumn{2}{c}{\textbf{CWS+POS}} \\ \hline
Model            & DCNN            & TE           & DCNN            & TE           & DCNN              & TE            \\ \hline
CWS (Block F1)    & \textbf{0.9738}          & 0.9683          & -               & -               & 0.9460            & \textbf{0.9503}           \\
POS (tagging acc) & -               & -               & 0.9056          & \textbf{0.9199}          & 0.9014            & \textbf{0.9097}           \\ \hline\hline
\end{tabular}%
}
\caption{Auxiliary-Module Experiment Results}
\label{tab:auxiliary_module}
\end{table}

\subsubsection{Evaluation of the Unified Front-end}
\label{subsubsec:evaluation_of_shooter}
We fine-tuned the unified front-end with three auxiliary module settings in Section~\ref{subsubsec:auxiliary_module}. In comparison, we used components from the pipeline-based front-end as baselines, which achieve SOTA performance in prosody prediction and G2P conversion respectively. The prosody baseline is a sequence labeling model with BERT boosting, and the G2P baseline is a BERT+CNN model. All the models were trained and tested using the People's Daily corpus described in Section~\ref{subsec:dataset}.

One can be seen from Table~\ref{tab:shooter_evaluation} that the auxiliary module is essential to our front-end's performance, which proves the effect of CWS and POS representations. Compared to baselines, we found that our front-end with POS and with CWS+POS auxiliary module got slightly improvement in polyphone disambiguation. As for prosody prediction, our front-end with CWS+POS auxiliary module obtained similar performance in PW prediction, and our front-end with any auxiliary module achieved significantly higher F1 score in IP prediction. However, the prosody baseline still performs the best in PP prediction. According to the test results, our front-end with CWS+POS auxiliary module can be regarded to have comparable performance with the baselines.

\begin{table}[]
\centering
\resizebox{\columnwidth}{!}{%
\begin{tabular}{l|cccc}
\hline\hline
\multicolumn{1}{c|}{\multirow{2}{*}{\textbf{Settings}}} & \multirow{2}{*}{\textbf{\begin{tabular}[c]{@{}c@{}}G2P \\ accuracy(\%) \end{tabular}}} & \multicolumn{3}{c}{\textbf{\begin{tabular}[c]{@{}c@{}}Prosody \\F1 score\end{tabular}}} \\
\multicolumn{1}{c|}{}                                   &                                                                                                             & \textbf{PW}                  & \textbf{PP}                  & \textbf{IP}                  \\ \hline
G2P baseline                                            & 95.92                                                                                                      & -                            & -                            & -                            \\
Prosody baseline                                        & -                                                                                                           & 0.9497                       & 0.8089                       & 0.8719                       \\ \hline
Ours (no auxiliary)                                       & 85.14                                                                                                      & 0.8795                       & 0.6670                       & 0.7085                       \\
Ours (CWS auxiliary)                                      & 95.07                                                                                                      & 0.9471                       & 0.7819                       & \textbf{0.9259}                       \\
Ours (POS auxiliary)                                      & \textbf{96.77}                                                                                                      & 0.8977                       & 0.8043                       & \textbf{0.9384}                       \\
Ours (CWS+POS auxiliary)                                  & \textbf{96.56}                                                                                                     & \textbf{0.9517}                       & 0.7745                       & \textbf{0.9457}                       \\ \hline\hline
\end{tabular}%
}
\caption{The G2P accuracy represents the polyphone disambiguation accuracy. The prosody F1 score in PP, PW and IP indicates the stacked prosody tagging F1 score.}
\label{tab:shooter_evaluation}
\end{table}
\subsubsection{Use SAR in the Unified Front-end}
In order to verify the effect of SAR, we did experiments using SAR in both training and evaluation stage and compared with scheduled sampling and auto-regressive (AR) methods. Table~\ref{tab:SAR} shows that despite the training methods, using SAR in evaluation can improve G2P accuracy significantly and make positive effects on prosody prediction as well when compared to the AR method. However, applying SAR in training severely degrades our front-end's performance. This is likely because for standalone polyphones, SAR always makes the polyphone step to see correct phoneme inputs from the previous step, and when there are an increasing number of continuous polyphones in the test dataset, our front-end may generate incorrect phoneme if the phoneme from the previous step is already incorrect.

\begin{table}[]
\centering
\resizebox{\columnwidth}{!}{%
\begin{tabular}{c|c|cccc}
\hline\hline
\textbf{Training}                   & \textbf{Evaluation} & \textbf{\begin{tabular}[c]{@{}c@{}}G2P\\ accuracy(\%)\end{tabular}} & \multicolumn{3}{c}{\textbf{\begin{tabular}[c]{@{}c@{}}Prosody\\F1 score\end{tabular}}} \\
\textbf{}                           & \textbf{}           & \textbf{}                                                         & \textbf{PW}                  & \textbf{PP}                  & \textbf{IP}                 \\ \hline
\multirow{2}{*}{Scheduled Sampling} & AR                  & 93.38                                                            & 0.9423                       & 0.7699                       & 0.9334                      \\
                                    & SAR                 & \textbf{96.56}                                                            & \textbf{0.9517}                       & \textbf{0.7745}                       & \textbf{0.9457 }                     \\ \hline
\multirow{2}{*}{SAR}                & AR                  & 70.41                                                            & 0.8818                       & 0.6721                       & 0.8954                      \\
                                    & SAR                 & \textbf{85.92}                                                            & \textbf{0.8828}                       & \textbf{0.6733}                      & \textbf{0.8955}                     \\ \hline\hline
\end{tabular}%
}
\caption{Semi-Auto-Regressive Experiment Results.}
\label{tab:SAR}
\end{table}

\subsubsection{Cascade with TTS Back-end}
\label{label:use_shooter_in_tts}
\begin{table}[]
\centering
\begin{tabular}{lc}
\hline\hline
\textbf{System}        & \textbf{MOS} \\ \hline
Ground Truth & $4.49\pm0.20$    \\
P-system  & $4.37\pm0.27$         \\
U-system  & $4.38\pm0.27$         \\ \hline\hline
\end{tabular}%
\caption{Mean option scores(MOS) with 95\% confidence intervals. Ground truth represents the human recordings. P-system is the one with the pipeline-based front-end, and U-system is the other one with our unified front-end.}
\label{tab:mos}
\end{table}
We implemented two front-end models, the pipeline-based and the unified (CWS+POS auxiliary) with Tacotron~\cite{wang2017tacotron} and WaveRNN~\cite{lafferty2001conditional} in two Mandarin TTS systems separately. In the training stage, Tacotron was trained using linguistic features from the corresponding front-end. In the synthesis stage, all the utterances were normalized by TN~\cite{zhang2020nntn} in advance. Table~\ref{tab:mos} shows that U-system can got comparable MOS with P-system, which is also close to the human recordings.

\section{Conclusion}
\label{sec:conclusion}
In this paper, we proposed a unified sequence-to-sequence front-end model for Mandarin TTS, to significantly simplify the Mandarin TTS front-end workflow. We also introduced a semi-auto-regressive inference method to improve the performance of our unified front-end. The experiment results show that our unified front-end can achieve comparable or better performance than the pipeline-based front-end in G2P and most prosody prediction tasks. Implementing our unified front-end in a Mandarin TTS system with Tacotron and WaveRNN can synthesize close-to-human speech with 4.38 MOS, and get comparable performance with using the pipeline-based frontend.

The potential future work includes using BERT to replace current character embedding and applying the unified front-end to other languages. Additionally, how to apply a flexible badcase-fixing schedule and improve the controllability of the unified front-end also need further investigations.

\vfill\pagebreak

\bibliographystyle{IEEEbib}

\begin{thebibliography}{10}

\bibitem{wang2017tacotron}
Yuxuan Wang, RJ~Skerry-Ryan, Daisy Stanton, Yonghui Wu, Ron~J Weiss, Navdeep
  Jaitly, Zongheng Yang, Ying Xiao, Zhifeng Chen, Samy Bengio, et~al.,
\newblock ``Tacotron: Towards end-to-end speech synthesis,''
\newblock {\em arXiv preprint arXiv:1703.10135}, 2017.

\bibitem{kalchbrenner2018efficient}
Nal Kalchbrenner, Erich Elsen, Karen Simonyan, Seb Noury, Norman Casagrande,
  Edward Lockhart, Florian Stimberg, Aaron van~den Oord, Sander Dieleman, and
  Koray Kavukcuoglu,
\newblock ``Efficient neural audio synthesis,''
\newblock {\em arXiv preprint arXiv:1802.08435}, 2018.

\bibitem{zhang2020nntn}
Junhui Zhang, Junjie Pan, Xiang Yin, Shichao Liu, Yang Zhang, Yuxuan Wang, and
  Zejun Ma,
\newblock ``A hybrid text normalization system using multi-head self-attention
  for mandarin,''
\newblock in {\em To Be Submitted to 2020 ICASSP}.

\bibitem{teahan2000compression}
William~J Teahan, Yingying Wen, Rodger McNab, and Ian~H Witten,
\newblock ``A compression-based algorithm for chinese word segmentation,''
\newblock {\em Computational Linguistics}, vol. 26, no. 3, pp. 375--393, 2000.

\bibitem{ying2001rnn}
Zhiwei Ying and Xiaohua Shi,
\newblock ``An rnn-based algorithm to detect prosodic phrase for chinese tts,''
\newblock in {\em 2001 IEEE International Conference on Acoustics, Speech, and
  Signal Processing. Proceedings (Cat. No. 01CH37221)}. IEEE, 2001, vol.~2, pp.
  809--812.

\bibitem{lu2019implementing}
Yanfeng Lu, Minghui Dong, and Ying Chen,
\newblock ``Implementing prosodic phrasing in chinese end-to-end speech
  synthesis,''
\newblock in {\em ICASSP 2019-2019 IEEE International Conference on Acoustics,
  Speech and Signal Processing (ICASSP)}. IEEE, 2019, pp. 7050--7054.

\bibitem{qian2010automatic}
Yao Qian, Zhizheng Wu, Xuezhe Ma, and Frank Soong,
\newblock ``Automatic prosody prediction and detection with conditional random
  field (crf) models,''
\newblock in {\em 2010 7th International Symposium on Chinese Spoken Language
  Processing}. IEEE, 2010, pp. 135--138.

\bibitem{pan2019mandarin}
Huashan Pan, Xiulin Li, and Zhiqiang Huang,
\newblock ``A mandarin prosodic boundary prediction model based on multi-task
  learning,''
\newblock {\em Proc. Interspeech 2019}, pp. 4485--4488, 2019.

\bibitem{xu2004grapheme}
Jun Xu, Guohong Fu, and Haizhou Li,
\newblock ``Grapheme-to-phoneme conversion for chinese text-to-speech,''
\newblock in {\em Eighth International Conference on Spoken Language
  Processing}, 2004.

\bibitem{mao2007inequality}
Xinnian Mao, Yuan Dong, Jinyu Han, Dezhi Huang, and Haila Wang,
\newblock ``Inequality maximum entropy classifier with character features for
  polyphone disambiguation in mandarin tts systems,''
\newblock in {\em 2007 IEEE International Conference on Acoustics, Speech and
  Signal Processing-ICASSP'07}. IEEE, 2007, vol.~4, pp. IV--705.

\bibitem{huang2008disambiguating}
Feng-Long Huang,
\newblock ``Disambiguating effectively chinese polyphonic ambiguity based on
  unify approach,''
\newblock in {\em 2008 International Conference on Machine Learning and
  Cybernetics}. IEEE, 2008, vol.~6, pp. 3242--3246.

\bibitem{vaswani2017attention}
Ashish Vaswani, Noam Shazeer, Niki Parmar, Jakob Uszkoreit, Llion Jones,
  Aidan~N Gomez, {\L}ukasz Kaiser, and Illia Polosukhin,
\newblock ``Attention is all you need,''
\newblock in {\em Advances in neural information processing systems}, 2017, pp.
  5998--6008.

\bibitem{skerry2018towards}
RJ~Skerry-Ryan, Eric Battenberg, Ying Xiao, Yuxuan Wang, Daisy Stanton, Joel
  Shor, Ron~J Weiss, Rob Clark, and Rif~A Saurous,
\newblock ``Towards end-to-end prosody transfer for expressive speech synthesis
  with tacotron,''
\newblock {\em arXiv preprint arXiv:1803.09047}, 2018.

\bibitem{shen2018natural}
Jonathan Shen, Ruoming Pang, Ron~J Weiss, Mike Schuster, Navdeep Jaitly,
  Zongheng Yang, Zhifeng Chen, Yu~Zhang, Yuxuan Wang, Rj~Skerrv-Ryan, et~al.,
\newblock ``Natural tts synthesis by conditioning wavenet on mel spectrogram
  predictions,''
\newblock in {\em 2018 IEEE International Conference on Acoustics, Speech and
  Signal Processing (ICASSP)}. IEEE, 2018, pp. 4779--4783.

\bibitem{mikolov2013distributed}
Tomas Mikolov, Ilya Sutskever, Kai Chen, Greg~S Corrado, and Jeff Dean,
\newblock ``Distributed representations of words and phrases and their
  compositionality,''
\newblock in {\em Advances in neural information processing systems}, 2013, pp.
  3111--3119.

\bibitem{shaheen2016impact}
Fatma Shaheen, Brijesh Verma, and Md~Asafuddoula,
\newblock ``Impact of automatic feature extraction in deep learning
  architecture,''
\newblock in {\em 2016 International Conference on Digital Image Computing:
  Techniques and Applications (DICTA)}. IEEE, 2016, pp. 1--8.

\bibitem{uszkoreit2017transformer}
Jakob Uszkoreit,
\newblock ``Transformer: A novel neural network architecture for language
  understanding,''
\newblock {\em Google Research Blog}, 2017.

\bibitem{wang2018style}
Yuxuan Wang, Daisy Stanton, Yu~Zhang, RJ~Skerry-Ryan, Eric Battenberg, Joel
  Shor, Ying Xiao, Fei Ren, Ye~Jia, and Rif~A Saurous,
\newblock ``Style tokens: Unsupervised style modeling, control and transfer in
  end-to-end speech synthesis,''
\newblock {\em arXiv preprint arXiv:1803.09017}, 2018.

\bibitem{battenberg2019effective}
Eric Battenberg, Soroosh Mariooryad, Daisy Stanton, RJ~Skerry-Ryan, Matt
  Shannon, David Kao, and Tom Bagby,
\newblock ``Effective use of variational embedding capacity in expressive
  end-to-end speech synthesis,''
\newblock {\em arXiv preprint arXiv:1906.03402}, 2019.

\bibitem{lafferty2001conditional}
John Lafferty, Andrew McCallum, and Fernando~CN Pereira,
\newblock ``Conditional random fields: Probabilistic models for segmenting and
  labeling sequence data,''
\newblock 2001.

\end{thebibliography}

\end{document}